\documentclass[conference]{IEEEtran}
\IEEEoverridecommandlockouts
\usepackage{microtype}
\usepackage{cite}
\usepackage{hyperref}
\usepackage{amsmath,amssymb,amsfonts}
\usepackage{algorithmic}
\usepackage{graphicx}
\usepackage{textcomp}
\usepackage{xcolor}
\usepackage{multicol}
\usepackage{parskip}
\usepackage{titlesec}
\usepackage{indentfirst}  
\usepackage{enumitem}

\titlespacing*{\section}{0pt}{0.5\baselineskip}{3pt}
\titlespacing*{\subsection}{0pt}{0.5\baselineskip}{2pt}
\begin{document}

\title{ORAssistant: A Custom RAG-based Conversational Assistant for OpenROAD}
\author{
\IEEEauthorblockN{Aviral Kaintura\IEEEauthorrefmark{1}\textsuperscript{§},
Palaniappan R\IEEEauthorrefmark{2}\textsuperscript{§},
Shui Song Luar\IEEEauthorrefmark{3},
Indira Iyer Almeida\IEEEauthorrefmark{3}}
\IEEEauthorblockA{\IEEEauthorrefmark{1}National Forensic Sciences University Delhi Campus, New Delhi, India\\
\IEEEauthorrefmark{2}BITS Pilani Hyderabad Campus, Hyderabad, Telangana, India\\
\IEEEauthorrefmark{3}Precision Innovations Inc.}
\IEEEauthorblockA{\IEEEauthorrefmark{1}102ctbmti2122019@nfsu.ac.in,
\IEEEauthorrefmark{2}f20212915@hyderabad.bits-pilani.ac.in,
\IEEEauthorrefmark{3}jluar@precisioninno.com,
\IEEEauthorrefmark{3}iiyer@precisioninno.com}
\vspace{-1.5em}
\thanks{\textsuperscript{§}Palaniappan R and Aviral Kaintura contributed equally to this project under the mentorship of Indira Iyer and Jack Luar.}
}

\maketitle

\begin{abstract}
Open-source Electronic Design Automation (EDA) tools are rapidly transforming chip design by addressing key barriers of commercial EDA tools such as complexity, costs, and access. Recent advancements in Large Language Models (LLMs) have further enhanced efficiency in chip design by providing user assistance across a range of tasks like setup, decision-making, and flow automation. This paper introduces ORAssistant, a conversational assistant for OpenROAD, based on Retrieval-Augmented Generation (RAG). ORAssistant aims to improve the user experience for the OpenROAD flow, from RTL-GDSII by providing context-specific responses to common user queries, including installation, command usage, flow setup, and execution, in prose format. Currently, ORAssistant integrates OpenROAD, OpenROAD-flow-scripts, Yosys, OpenSTA, and KLayout. The data model is built from publicly available documentation and GitHub resources. The proposed architecture is scalable, supporting extensions to other open-source tools, operating modes, and LLM models. We use Google Gemini as the base LLM model to build and test ORAssistant. Early evaluation results of the RAG-based model show notable improvements in performance and accuracy compared to non-fine-tuned LLMs. 
\end{abstract}


\section{Introduction}
Since 2018, open-source EDA tools have rapidly democratized hardware design and driven innovation through research and collaboration. Free from licensing constraints, they foster a thriving ecosystem of chip design and education\cite{kahng2020open,efabless_chipignite,venn2024tiny}. 
Recently, ML and GenAI-based chip design methodologies have been applied to open-source tools, yielding significant benefits in productivity \cite{efabless_genai_video}. The open infrastructure of these tools simplifies model training, integration and enable the use of shared resources, such as documentation, scripts, and datasets for a host of deployment options\cite{aws_blog}.

ORAssistant began as a Google Summer of Code (GSoC) 2024 \cite{osre2024} project\textsuperscript{§} with an objective of assisting OpenROAD users in completing basic tasks successfully — from setup to flow execution. We focused on addressing frequently occurring problems in areas such as installation, design setup and flow, command usage, and debugging. Traditional user resources, such as documentation and tutorials, tend to become outdated quickly and fail to incorporate practical knowledge gained from collaborative user experiences. Our goal was to build a chatbot that harnesses the dynamic nature of open-source tools and resources to address basic OpenROAD tasks efficiently. We developed a publicly available chatbot\cite{orassistant_frontend} that supports continuous improvement and scalability across the tool chain. The complete source code for ORAssistant is available on GitHub \cite{orassistant_github}.

ORAssistant's RAG architecture can be used atop any publicly available LLM. The RAG system enhances the base LLM's output by ensuring that knowledge is retrieved from trusted data sources, generating reliable responses \cite{lewis2020retrieval}. Our key contributions are:

\begin{itemize}[topsep=0pt]
    \item \textbf{Development of ORAssistant}: A conversational AI agent that assists users in a simple question and answer format with basic conversational abilities.
    \item \textbf{ORAssistant RAG Dataset}: A curated dataset derived from open-source tool documentation and \textsc{GitHub} data from the OpenROAD and OpenROAD-flow-scripts repositories. 
    \item \textbf{ORAssistant Evaluation Framework}: Development of a comprehensive evaluation methodology that utilizes both the publicly available EDA Corpus dataset \cite{wu2024eda} and a self-curated question-answer (QA) dataset. This framework enables robust assessment of LLM performance in the OpenROAD domain, facilitating comparisons between ORAssistant and non-fine-tuned LLMs.
\end{itemize}

\section{Dataset Generation}
The effectiveness of any RAG-based system depends on the quality of the underlying data sources, as they form the basis for model accuracy, information retrieval, and response generation. Thus, curating a properly annotated dataset is crucial. 

\subsection{Data Sources}
ORAssistant primarily assists users of OpenROAD and OpenROAD-flow-scripts while also offering basic support across the RTL-to-GDSII tool chain, from synthesis to layout verification. Its knowledge base includes public documentation, tool manuals, and custom-annotated \textsc{GitHub} discussions, as shown below:
\begin{itemize}[topsep=0pt]
    \item OpenROAD Documentation \cite{openroad_docs}
    \item OpenROAD-flow-scripts Documentation \cite{openroad_flow_scripts_docs}
    \item OpenROAD Man pages \cite{openroad_docs}
    \item OpenSTA Documentation \cite{opensta_docs}
    \item Yosys Documentation \cite{yosys_docs}
    \item KLayout Documentation \cite{klayout_docs}
    \item Research Papers, Tutorials on OpenROAD \cite{openroad_project}
    \item Reformatted and labeled \textsc{GitHub} discussions \cite{openroad_repo} 
\end{itemize}

These diverse sources capture detailed information about the tools and their usage. To keep the dataset updated, an automated build script is used to extract and version data from the OpenROAD repository and external sources, ensuring that ORAssistant is equipped with the most relevant and up-to-date information. Additionally, live hyperlinks for each data source are stored, enabling citations during response generation.

\subsection{Issues and Discussion Analysis}
Along with public documentation sources,  selectively curated conversations from support forums such as \textsc{GitHub} issues and discussions have been incorporated to identify key problem areas. Using the \textsc{GitHub} \textsc{GraphQL API},  conversations from the issues and discussions pages of the OpenROAD and OpenROAD-flow-scripts repositories, are extracted and stored in \textsc{JSONL} datasets. Since these scraped conversations lack proper annotation, an LLM-based categorization approach has been used to generate three tags for each conversation: category, subcategory, and referenced tools. This characterization enables the correct routing of the tool-based RAG system to domain-specific retriever tools. Detailed analysis indicated that GitHub Issues were predominantly bug reports with limited relevance. GitHub Discussions provided a wider variety of user queries and their corresponding solutions, making it a more valuable data source for ORAssistant.

The current \textsc{JSONL} dataset \cite{orassistant_rag_dataset} consists of a total of 736 issues and 344 discussions.
Table \ref{tab:issue_discussion_distribution} summarizes the distribution of issues and discussions by category.

\begin{table}[ht]
\caption{Distribution of GitHub Issues and Discussions by Category}
\begin{center}
\begin{tabular}{|l|c|c|}
\hline
\textbf{Category} & \textbf{Issues (\%)} & \textbf{Discussions (\%)} \\
\hline
Bug & 45.90 & 4.65 \\
Feature request & 18.60 & 8.72 \\
Runtime & 13.60 & 28.50 \\
Build & 9.92 & 8.14 \\
Query & 7.34 & 40.70 \\
Installation & 2.85 & 3.49 \\
Documentation & 0.95 & 1.16 \\
Configuration & 0.82 & 4.65 \\
\hline
\end{tabular}
\end{center}
\label{tab:issue_discussion_distribution}
\end{table}

\section{RAG Architecture}
ORAssistant's tool-based RAG architecture and data ingestion pipeline can be integrated with any LLM, supporting both local and cloud-based deployments. This allows organizations to balance performance requirements, computational resources, and data privacy concerns.
\subsection{Dataset Ingestion}
The pipeline processes documents in multiple formats (Markdown, PDF, HTML), and divides them into smaller, manageable document chunks based on format-specific delimiters. These chunks are then fed to a SBERT-based \cite{reimers-2019-sentence-bert} embedding model, which encodes the textual information into dense vector representations. The resulting vectors are then stored in a Facebook AI Similarity Search (FAISS) \cite{douze2024faiss} vector database for downstream use. While this approach efficiently captures the semantic meaning in each chunk, it is unsuitable for retrieving documents based on exact keywords. To perform exact term matching, we use the classical Best Match 25 (BM25) \cite{stuart2010matching} indexing technique, which allows for keyword-based retrieval. The knowledge base is thus represented as a weighted sum of vectors and keyword indices, allowing for both semantic and exact-term matching. 
\begin{figure}[htbp]
\centerline{\includegraphics[width=8.5 cm ]{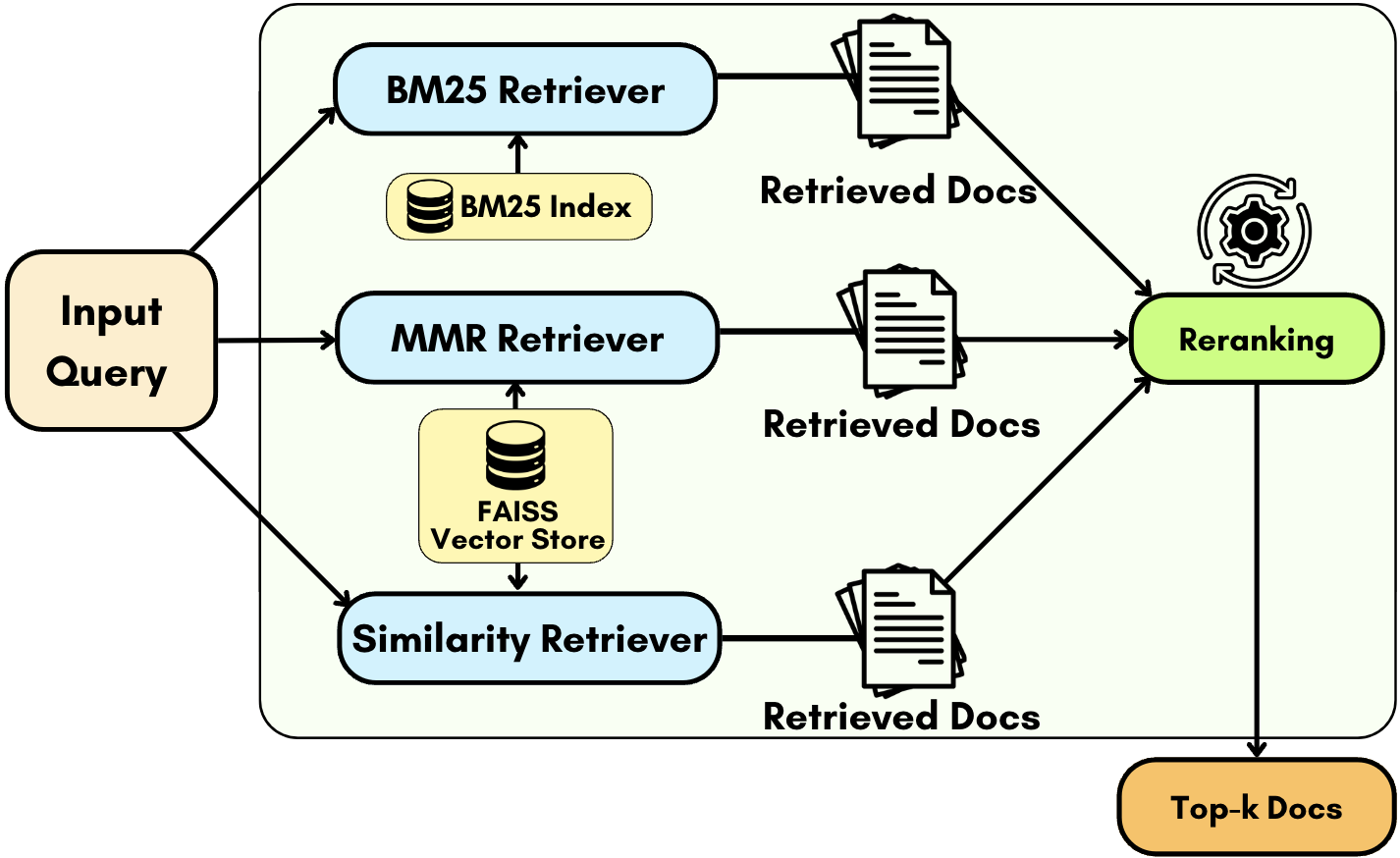}}
\caption{Hybrid Retriever Function}
\label{fig:Retrievers.png}
\end{figure}
\subsection{Hybrid Retriever Function}
As illustrated in Figure \ref{fig:Retrievers.png}, the retriever function searches its input knowledge base to identify document chunks most relevant to a given query. It employs multiple vector search techniques, including similarity search and maximal marginal relevance (MMR) search. In similarity search, chunks with the highest cosine similarity to the input are selected, while MMR search introduces diversity by minimizing redundancy in the retrieved chunks. Additionally, a classical text-based search is performed on the BM25 index to retrieve documents containing the exact keywords specified in the query. A re-ranking model then processes these search results, adjusting the document ranking to provide a top-$k$ selection of the most relevant documents, ensuring both precision and diversity in the final output.

\begin{figure}[htbp]
\centering
\includegraphics[width=7.5cm]{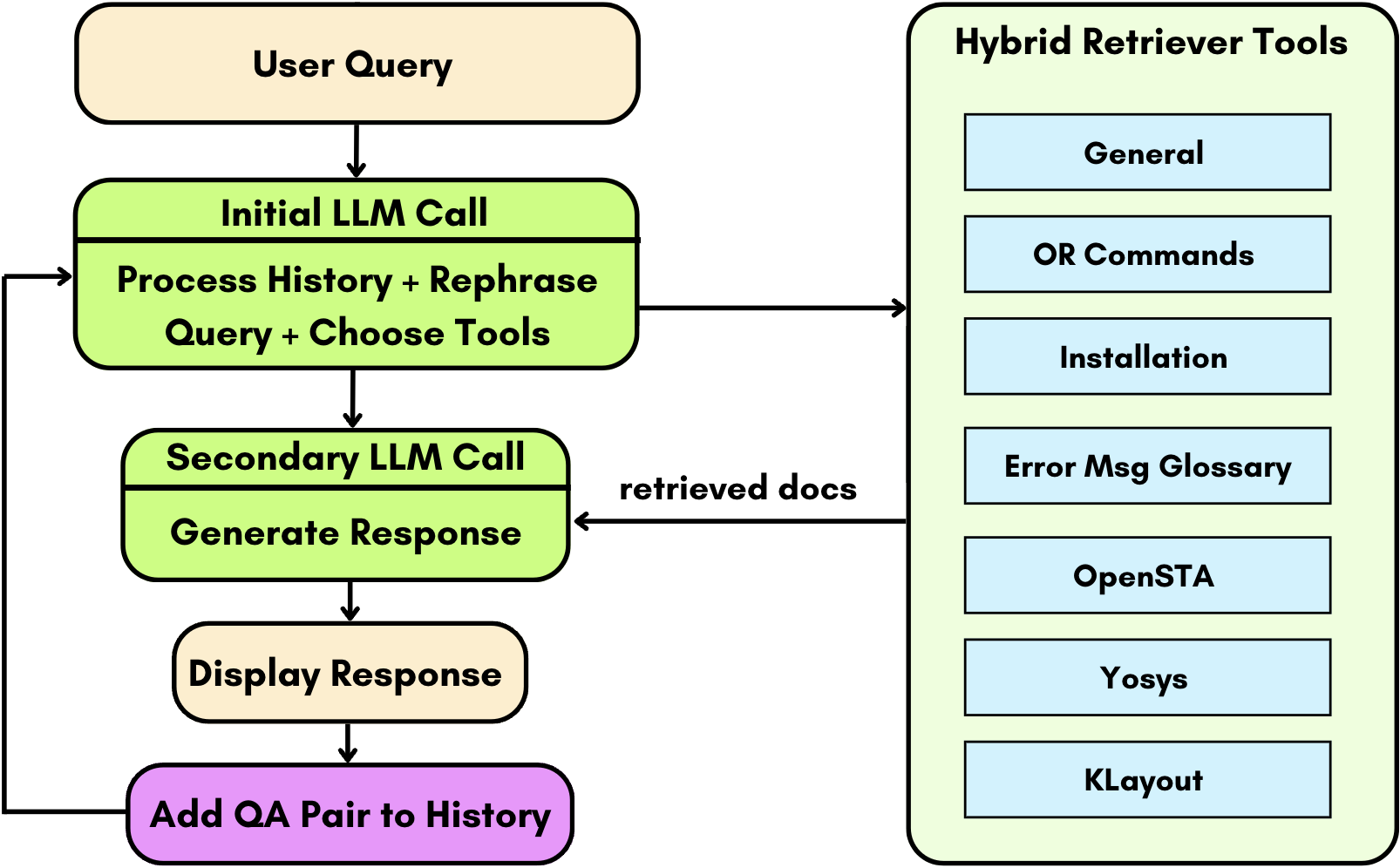}
\caption{ORAssistant's tool-based RAG Architecture}
\label{fig:architecture}
\end{figure}

\subsection{Domain Specific Retriever Tools}
ORAssistant's knowledge base encompasses a wide range of information from various applications in the OpenROAD flow. Subsets of this knowledge base relevant to specific applications are provided to the hybrid retriever function to form domain specific retriever tools, as listed in Figure \ref{fig:architecture}. A custom prompt guides the base LLM in selecting the appropriate retriever tools for each query. For instance, the \textit{OR Commands} tool retrieves information specific to the OpenROAD framework's commands, while the \textit{Installation} tool focuses on documentation related to installation procedures. In contrast to using a single retriever function on a pooled knowledge base, the modular approach significantly reduces the chances of incorrect document retrieval. Additionally, the architecture allows for future integration with other open-source tools and flow runners within the OpenROAD ecosystem. 

\subsection{Context-Aware Response Generation}
As the conversation progresses, ORAssistant stores question-answer pairs locally to maintain context. When the user submits a new query, the system first processes the stored conversation history to ensure context continuity. The incoming query is then rephrased, incorporating information from previous exchanges. This gives ORAssistant the capability to answer follow up queries, and thereby maintain long, context-aware conversations.

The tool-based RAG architecture ensures that responses are context-correct by leveraging both the conversation history and domain-specific knowledge sources. These sources enable the system to provide responses with precise citations and hyperlinks for each query. Figure \ref{fig:architecture} depicts the tool-based RAG system operating in two distinct stages:
\begin{itemize}
    \item \textbf{Initial LLM Call}: ORAssistant processes the conversation history to contextualize and rephrase the user's current query. Based on the refined query, the most appropriate tools are selected and documents are subsequently retrieved.
    \item \textbf{Secondary LLM Call}: ORAssistant generates a response for the rephrased query using the documents retrieved by the selected tools.
\end{itemize}

\section{Hosted Application}
\begin{figure}[htbp]
\centering
\includegraphics[width=9.5cm]{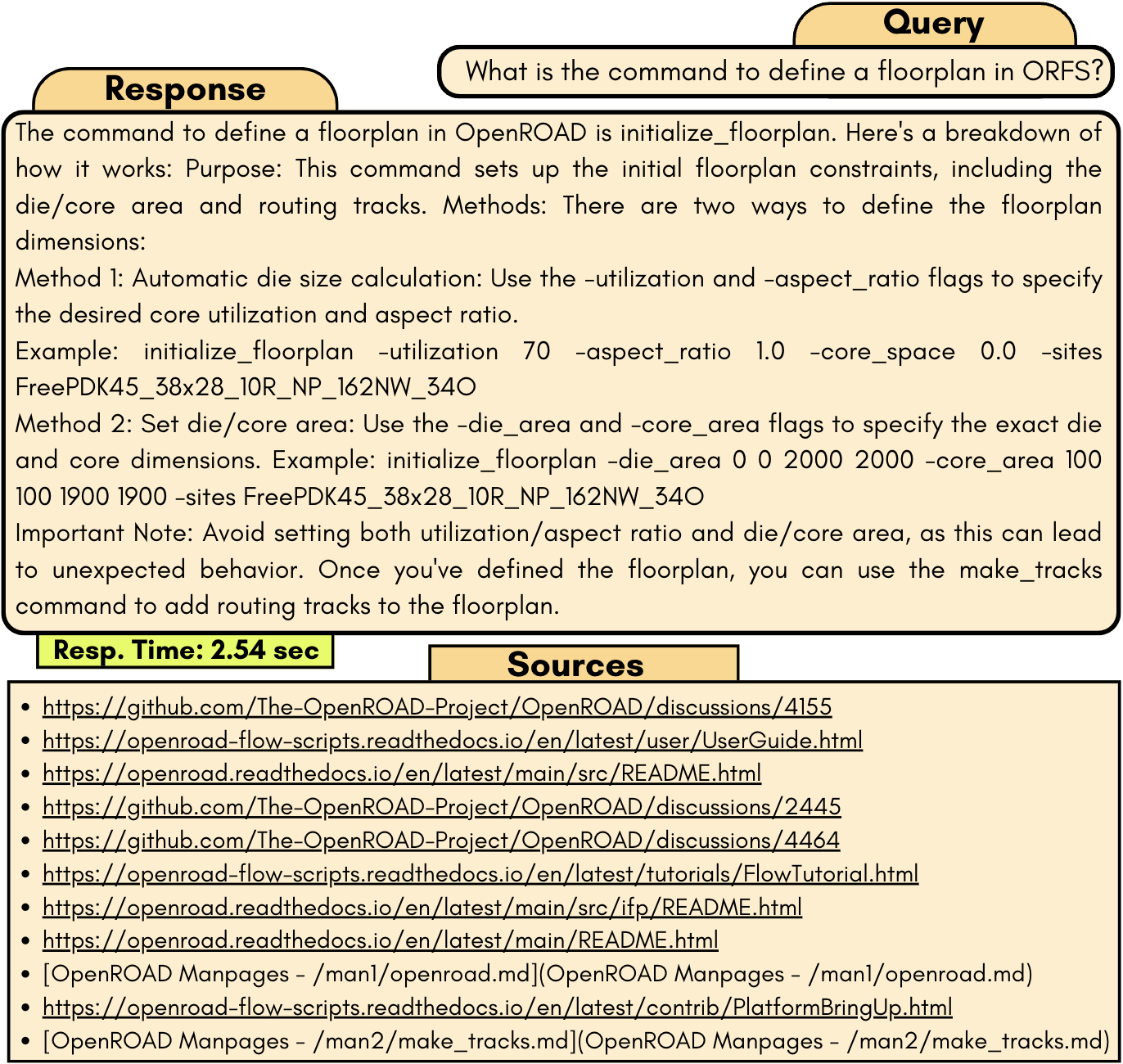}
\caption{Example of ORAssistant generated response.}
\label{fig:example_response}
\end{figure}

ORAssistant can be accessed on a \textsc{Next.js}\cite{nextjs} based web front-end\cite{orassistant_frontend}. The hosted version uses Gemini 1.5 Flash \cite{gemini_flash} as its base LLM alongside the \textit{text-embedding-004}\cite{google_text_embeddings} model for generating vector embeddings. The web application supports the creation of multiple conversation threads, each retaining its own history. This allows users to switch between threads while keeping each discussion focused and relevant to specific topics. Figure \ref{fig:example_response} shows an ORAssistant-generated, composite response for a user query about floorplan creation options in OpenROAD-flow-scripts. 

\section{Evaluation and Results}
    
The evaluation of ORAssistant is crucial to understanding its performance and limitations. The process aims to quantify the system's ability to successfully retrieve information from correct sources and provide precise responses. The evaluation process identifies areas where the model struggles, allowing for continuous improvement. Additionally, it helps identify gaps in the original documentation and knowledge sources. These insights can guide future improvements through a bidirectional feedback loop.

To assess the effectiveness of our approach, we compared ORAssistant's performance against base pre-trained LLMs like Gemini 1.5 Flash \cite{gemini_flash} and GPT-4o \cite{gpt4o}. We utilize an approach based on GPTScore \cite{fu2023gptscore}, where a separate LLM is used as an automated evaluator. This \textit{LLM Judge} is given the original question, a ground truth answer, and the model-generated response. Using a carefully designed system prompt, the \textit{LLM Judge} then assesses the quality, coherence, and accuracy of the generated response by comparing it to the ground truth. The evaluation generates the following metrics:
\begin{itemize}
    \item \textbf{Classification metrics}: As shown in Table \ref{tab:evalmetrics}, the judge compares the ground truth and LLM's response, classifying them into one of four predefined categories (TP, TN, FP, FN). Metrics such as Accuracy, Precision, Recall, and F1 score are then computed using these classifications.
    \item \textbf{LLMScore}: The judge assigns a score in the range of $[0,1]$, based on the quality and accuracy of the LLM's response in relation to the ground truth.
\end{itemize}
\begin{table}[ht]
    \caption{Evaluation Metrics for Model Answers}
    \resizebox{\columnwidth}{!}{
    \begin{tabular}{|p{8.0cm}|}
    \hline
    \textbf{Q:} What does CTS stand for? \newline \textbf{A:} CTS stands for Clock Tree Synthesis. It is a stage... \newline \textbf{Eval:} True Positive (TP) (Detailed, accurate, and relevant.) \\ \hline
    \textbf{Q:} What is the latest movie released? \newline \textbf{A:} I can't provide information on movies... \newline \textbf{Eval:} True Negative (TN) (Correctly identified out of scope.) \\ \hline
    \textbf{Q:} What does CTS stand for? \newline \textbf{A:} CTS stands for Central Time Scheduling... \newline \textbf{Eval:} False Positive (FP) (Incorrect and irrelevant.) \\ \hline
    \textbf{Q:} What does CTS stand for? \newline \textbf{A:} I cannot provide an answer... \newline \textbf{Eval:} False Negative (FN) (Failed to answer when expected.) \\ \hline
    \end{tabular}
    }
    \label{tab:evalmetrics}
\end{table}

For our evaluation, we used two QA datasets for evaluation: a custom curated HumanEval dataset with 50 OpenROAD-related QA pairs \cite{orassistant_humaneval}, and 100 QA pairs from the publicly available EDA Corpus dataset\cite{wu2024eda}. 

To ensure a fair comparison, we conducted five independent runs for each question across these models: ORAssistant (with Gemini 1.5 Flash), base GPT-4o, and base Gemini 1.5 Flash. Multiple runs help account for the variability in LLM outputs, reducing outliers and improving statistical reliability.  Evaluation metrics computed using Gemini 1.5 Pro\cite{google_gemini_pro} as the judge LLM have been averaged for each dataset and presented in Table \ref{tab:evalresults}. As depicted in Table \ref{tab:evalresults}, ORAssistant significantly outperforms the base pre-trained LLMs across both the EDA Corpus and HumanEval datasets. ORAssistant achieves notably high precision and recall scores, indicating very few false positives and false negatives in its responses. In contrast, both GPT-4o and Gemini 1.5 Flash exhibit subpar performance. Although GPT-4o achieves high recall scores on both datasets, it is offset by a very low precision score. This suggests that the model often hallucinates and generates false positive responses, without acknowledging its lack of knowledge.  Across both datasets, ORAssistant attains a considerably high LLMScore, when compared to the base pre-trained LLMs. In terms of response times, ORAssistant averages 2.6 seconds across the testing datasets, while base GPT-4o records 4.7 seconds and base Gemini 1.5 Flash averages 2.3 seconds.
\begin{table}[ht]
\centering
\caption{Evaluation Results on EDA Corpus (100 Questions) and Human Eval (50 Questions) Datasets.}
\resizebox{0.5\textwidth}{!}{ 
\begin{tabular}{|l|c|c|c|c|c|}
\hline
\multicolumn{6}{|c|}{\textbf{EDA Corpus Dataset}} \\ \hline
Architecture & Acc. (\%) & Prec. (\%) & Rec. (\%) & F1 (\%) & LLMScore (\%) \\ \hline
ORAssistant & \textbf{90.4} & \textbf{94.8} & 95.2 &  \textbf{95.0} & \textbf{77.7} \\ \hline
GPT-4o & 48.4 & 48.4 & \textbf{100.0} & 65.2 & 52.6 \\ \hline
Gemini 1.5 Flash & 38.0 & 43.3 & 75.7 & 55.1 & 35.7 \\ \hline
\multicolumn{6}{|c|}{\textbf{Human Eval Dataset}} \\ \hline
Architecture & Acc. (\%) & Prec. (\%) & Rec. (\%) & F1 (\%) & LLMScore (\%) \\ \hline
ORAssistant & \textbf{87.2} & \textbf{92.4} & 94.0 & \textbf{93.2} & \textbf{79.7} \\ \hline
GPT-4o & 46.8 & 46.8 & \textbf{100.0} & 63.8 & 48.7 \\ \hline
Gemini 1.5 Flash & 32.8 & 35.4 & 80.2 & 49.1 & 28.1 \\ \hline
\end{tabular}
}
\label{tab:evalresults}
\end{table}
\vspace{1em}

Since ORAssistant uses Gemini 1.5 Flash, the contrast between its scores and the base Gemini 1.5 Flash scores highlights how the tool-based RAG architecture guides the LLM towards better performance. While base pre-trained models struggle with relevance due to general training data and outdated knowledge, ORAssistant enhances accuracy by leveraging up-to-date data sources. Moreover, ORAssistant avoids hallucinations by grounding its responses in reliable, deterministic, and contextually relevant data sources.

\section{Related Work}
 An alternate approach for an OpenROAD Assistant\cite{sharma2024openroad} as a chatbot and script generator, uses a fine-tuned LLM. Other tools like the Hybrid RAG based Ask-EDA \cite{shi2024askeda} and the domain-adaptive ChatNeMo \cite{liu2023chipnemo} use proprietary data. ChatEDA 
\cite{wu2024chateda}  employs fine-tuning for basic task planning in physical design using OpenROAD and other EDA tools. In \cite{pu2024customized}, the authors utilize a RAG based system tailored for OpenROAD and EDA tools, with custom fine-tuned embeddings and reranker models. 

RAG offers greater flexibility through real-time adaptation to tool and data source changes. Our tool-based architecture is scalable and aligns well with OpenROAD's modular flow.

\section{Future Work}
Our tool-based architecture enables support for interfacing with OpenROAD's Python-based APIs for custom applications. Combining a fine-tuned model with our RAG architecture will provide dual advantages of real-time adaptability for documentation and enhanced accuracy for tasks like design exploration and script generation. To further enhance the performance of the retriever functions, embeddings and reranker models can be fine-tuned on ORAssistant's knowledge base. Adding human-in-the-loop feedback is another way to continuously improve both the knowledge base and generated responses. We also plan to interface ORAssistant with OpenROAD command-line interface (CLI) and graphical-user-interface (GUI). 

\section{Conclusion}
In this paper, we present ORAssistant, a RAG-based assistant built using a scalable, tool-based architecture for the OpenROAD flow. This chatbot assists users by providing contextual and reliable answers for common user queries using native and publicly available data sources. Initial results show that the RAG model outperforms base pre-trained LLMs from metrics derived from human evaluations and automated LLM-based methods. ORAssistant showcases the potential of GenAI-based tools in chip design to enhance user experience, enabling users to learn faster and gain deeper insights across all levels of expertise. This work, supported by the OpenROAD project team, GSoC, and OSRE, is the result of collaborative and open-source community-driven EDA advocacy, and contribution.

\newpage
\bibliographystyle{ieeetrans}
\bibliography{biblo}

\end{document}